\documentclass[a4paper, 10pt, conference]{ieeeconf}

\usepackage{amsmath,amssymb,amsfonts,mathrsfs}

\usepackage{graphicx}
\usepackage[labelformat=simple]{subcaption} 
\usepackage{caption}
\usepackage{float}
\usepackage{dblfloatfix}
\usepackage[table,xcdraw]{xcolor}

\usepackage{booktabs}
\usepackage{longtable} 
\usepackage{tabularx}
\usepackage{multirow}
\usepackage{array}
\usepackage{bigstrut}

\let\labelindent\relax
\usepackage{enumitem}

\usepackage{algorithmic}

\usepackage{url}
\usepackage{textcomp}
\usepackage{cite}

\usepackage{pifont}

\usepackage{tikz}
\usepackage[framemethod=tikz]{mdframed}
\usepackage{balance}
\usepackage{afterpage}
\usepackage{placeins}
\usepackage{changepage}
\usepackage{listings}
\usepackage{comment}
\usepackage{cuted} 
\usepackage{breqn}  
\definecolor{myblue}{RGB}{220,230,250}
\definecolor{colborder}{RGB}{100,100,100}
\definecolor{lightgray}{gray}{0.95}

\def\BibTeX{{\rm B\kern-.05em{\sc i\kern-.025em b}\kern-.08em
    T\kern-.1667em\lower.7ex\hbox{E}\kern-.125emX}}

\makeatletter
\long\def\@IEEEtitleabstractindextextbox#1{\parbox{0.922\textwidth}{#1}}
\makeatother

\IEEEoverridecommandlockouts                            
\title{\LARGE \bf
Towards Zero-Shot Differential Morphing Attack Detection with Multimodal Large Language Models 
}

\author{%
    Ria Shekhawat$^{\dagger}$ \quad  
    Hailin Li$^{\dagger}$ \quad 
    Raghavendra Ramachandra$^{\dagger}$ \quad  
    Sushma Venkatesh$^{\ddagger}$ \\
    \textit{$^{\dagger}$Norwegian University of Science and Technology (NTNU), Gj{\o}vik, Norway.} \\
    \textit{$^{\ddagger}$MOBAI AS, Gj{\o}vik, Norway.} \\
    E-mail: 
		\{raghavendra.ramachandra\}@ntnu.no}

\begin{document}

\maketitle

\begin{abstract}
Leveraging the power of multimodal large language models (LLMs) offers a promising approach to enhancing the accuracy and interpretability of morphing attack detection (MAD), especially in real-world biometric applications. This work introduces the use of LLMs for differential morphing attack detection (D-MAD). To the best of our knowledge, this is the first study to employ multimodal LLMs to D-MAD using real biometric data. To effectively utilize these models, we design Chain-of-Thought (CoT)-based prompts to reduce failure-to-answer rates and enhance the reasoning behind decisions.
Our contributions include: (1) the first application of multimodal LLMs for D-MAD using real data subjects, (2) CoT-based prompt engineering to improve response reliability and explainability, (3) comprehensive qualitative and quantitative benchmarking of LLM performance using data from 54 individuals captured in passport enrollment scenarios, and (4) comparative analysis of two multimodal LLMs: ChatGPT-4o and Gemini providing insights into their morphing attack detection accuracy and decision transparency.
Experimental results show that ChatGPT-4o outperforms Gemini in detection accuracy, especially against GAN-based morphs, though both models struggle under challenging conditions. While Gemini offers more consistent explanations, ChatGPT-4o is more resilient but prone to a higher failure-to-answer rate.
\end{abstract}

\section{INTRODUCTION}
With rapidly evolving automation, Face Recognition Systems (FRS) have become essential for security and identity verification, commonly used in scenarios such as smartphone unlocking, passport issuance, and Automated Border Control (ABC) systems. However, with the rapid development of facial morphing techniques, these systems are increasingly susceptible to morphing attacks, as discussed in prior research \cite{8053499},\cite{venkatesh2021face}. These attacks involve generating synthetic facial images by merging features from two or more individuals, potentially enabling unauthorized access by multiple people through a single morphed image.

\begin{figure}[t!]
	\centering
	\includegraphics[width = 1\linewidth]{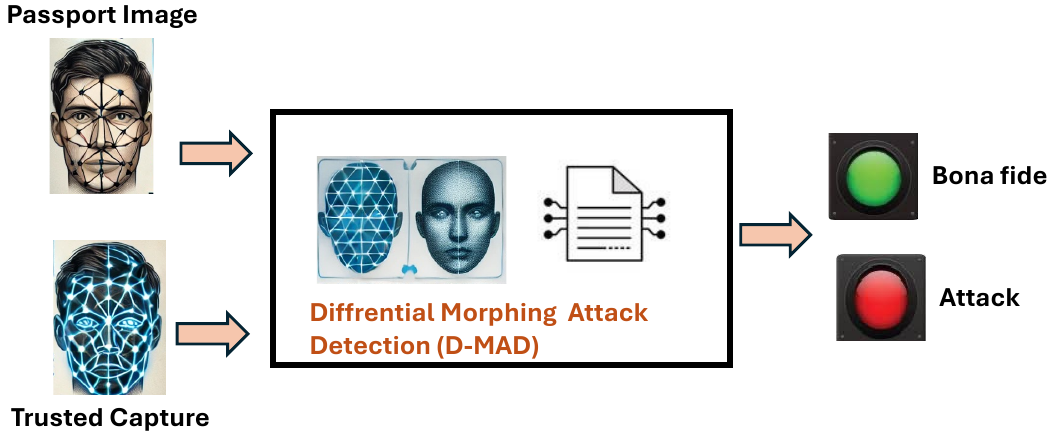}
	\caption{In a typical D-MAD scenario, facial images captured from a passport and an Automated Border Control (ABC) gate are used to extract facial features. These features are compared and analyzed to detect any signs of morphing.}
	\label{fig:Intro}
\end{figure}

Rising concerns about the security risks posed by morphing attacks have driven the creation of specialized methods known as Morphing Attack Detection (MAD). These techniques are generally divided into two main types \cite{venkatesh2021face}: Single-image MAD (S-MAD), which analyzes a single image for signs of manipulation, and Differential MAD (D-MAD), which compares two images typically a live capture and a stored reference to detect inconsistencies representing the morphing process.  Compared to S-MAD methods, D-MAD techniques have shown greater reliability and better generalization across various morphing generation methods. This improved performance is largely due to the use of a reference facial image, which provides a direct comparison and enhances the system’s ability to identify subtle cues of morphing. Figure \ref{fig:Intro} illustrates the D-MAD process, where two facial images, one from a passport and the other from a trusted source such as an Automated Border Control (ABC) gate that are provided for the same individual. These images are then analyzed using D-MAD techniques to determine whether the passport photo has been digitally morphed.

D-MAD techniques have been widely explored in research and are generally categorized into two main approaches: feature-based methods and de-morphing-based methods. Feature-based methods focus on extracting and comparing features from a pair of facial images—typically one from a passport and the other from a trusted source. These features, which can be derived using either handcrafted descriptors or deep learning models, are analyzed to detect discrepancies that may indicate morphing. On the other hand, de-morphing techniques aim to reverse the morphing process by reconstructing the original, hidden face from a morphed image. These methods are particularly effective in controlled environments where both the passport and reference images are of high quality. Like feature-based approaches, de-morphing can be implemented using traditional handcrafted techniques or modern deep learning frameworks.

The feature difference-based approach relies on extracting and comparing features from both images in a pair, using either handcrafted or deep learning methods. Handcrafted features include techniques such as Local Binary Patterns (LBP), Binarized Statistical Image Features (BSIF), multimodal and hybrid features \cite{ramachandra2022residual} \cite{ramachandra2023differential}, as well as wavelet-based features \cite{Aghdaie-waveletMAD-IJCB-2021}, SLERP features \cite{singh2022reliable}. These features are typically classified using linear Support Vector Machines (SVMs), as demonstrated in several studies \cite{Raghavendra-DetectingMorphedFace-BTAS-2016, Aghdaie-waveletMAD-IJCB-2021, Raghavendra-MADusingScaleSpaceFeature-ISBA-2019, raghavendra2022multimodality}. In addition, handcrafted 3D face information has been employed to capture depth-based differences between genuine and morphed images. This approach, which uses depth features to distinguish morphing artifacts, has shown promising results—particularly on the ABC dataset \cite{singh2022reliable, singh2022fusion}. Handcrafted features that focus on reversing traditional landmark-based morphing techniques are commonly utilized in de-morphing approaches. These methods aim to undo the morphing process and recover the original facial components, as demonstrated in several studies \cite{Ortega-BorderControlMAD-IEEEAcess-2020, zhang2023morphganformer, Ferrara-FaceDemorphing-IEEE-EUSIPCO-2018}.

Deep learning methods have become a prominent focus in the development of Differential Morphing Attack Detection (D-MAD) systems \cite{Raghavendra-DNNMorphingDetection-CVPR-2017, Damer-anomalyMAD-BTAS-2019}. These approaches include leveraging pre-trained neural networks for extracting facial features \cite{Raghavendra-DNNMorphingDetection-CVPR-2017}, adapting models through fine-tuning for domain-specific tasks \cite{Kraetzer-modellingMorphingAttacks-IHMMSec-2017}, and building fully end-to-end deep learning architectures \cite{seibold-neural-network-MAD-ISA-2020}. Features derived from models trained on large-scale datasets like ImageNet \cite{singh2022reliable, ramachandra-algorithmic-fairness-CVPR-2022} and face-centric datasets \cite{Scherhag-FaceMorphingAttacks-TIFS-2020} have demonstrated strong performance in detecting morphing attacks. Moreover, advanced architectures such as Siamese and Double Siamese networks \cite{Borghi-DoubleSiamese-mdpi-2021, soleymani-DMAD-Siamese-2021} have shown significant potential for accurately identifying morphed images in D-MAD settings.

Recent progress in foundation models and multimodal Large Language Models (LLMs) has opened new possibilities for applying these technologies to morphing attack detection. Several recent studies have explored their use in Single-image Morphing Attack Detection (S-MAD) \cite{MADCLIP}, \cite{zhang2024ViT}, \cite{caldeira2025madationfacemorphingattack}, \cite{zhang2025chatgptencountersmorphingattack}. The pioneering work in this area is presented in \cite{MADCLIP}, where a foundation model is used in a zero-shot setting, utilizing natural language prompts to assist in detecting morphing attacks. Building on this idea, subsequent research has proposed fine-tuned versions of various foundation models to improve detection performance \cite{zhang2024ViT}, \cite{caldeira2025madationfacemorphingattack}. The first study to introduce a multimodal LLM for S-MAD is detailed in \cite{zhang2025chatgptencountersmorphingattack}, where ChatGPT-4o is employed to detect morphing attacks. This approach not only outperformed traditional foundation models in accuracy but also provided natural language explanations for its decisions, adding a layer of interpretability to the detection process. The emerging use of multimodal large language models (LLMs) in biometric applications offers two key advantages over conventional approaches:
(a) the ability to generalize across different types of morphing techniques, and
(b) the capability to provide natural language explanations for their decisions.
However, multimodal LLMs are also prone to hallucinations and often fail to answer queries reliably. A common approach to mitigate these issues is to force the model to generate a response and to repeat the query multiple times, which can help reduce hallucinations and improve response consistency.

In this work, we introduce the use of multimodal large language models (LLMs) for differential morphing attack detection (D-MAD) applications. To the best of our knowledge, this is the first study to explore D-MAD using multimodal LLMs. Furthermore, it is also the first to evaluate such models on real biometric samples, rather than synthetic face images, with data collected following user consent—thereby providing insights into real-life deployment scenarios.

To effectively utilize multimodal LLMs, we designed prompts using the Chain-of-Thought (CoT) approach to reduce the rate of unanswered queries and to encourage more reliable reasoning behind the model's decisions.
The main contributions of this work are as follows:
\begin{itemize}
    \item First application of multimodal LLMs for differential morphing attack detection using real biometric data.
    \item Development of Chain-of-Thought (CoT) based prompting to improve answer reliability and enhance the reasoning behind decisions.
    \item Extensive experiments benchmarking both qualitative and quantitative performance of multimodal LLMs using data from 54 real data subjects captured in passport enrolment scenarios.
    \item Comparative analysis of morphing attack detection performance between two multimodal LLMs, ChatGPT-4o and Gemini, offering insights into decision accuracy and explainability.
\end{itemize}

The rest of the paper is organised as follows: Section \ref{sec:llm} describes the proposed framework using multimodal LLMs for D-MAD, Section \ref{sec:DB} presents the database used in this work, Section \ref{sec:results} discuss the quantitative and qualitative results of the multimodal LLMs for D-MAD and Section \ref{sec:conc} draws the conclusion. 

\section{Proposed Framework for D-MAD using Multimodal LLMs}
\label{sec:llm}
\begin{figure*}[htp]
	\centering
	\includegraphics[width = 1\linewidth]{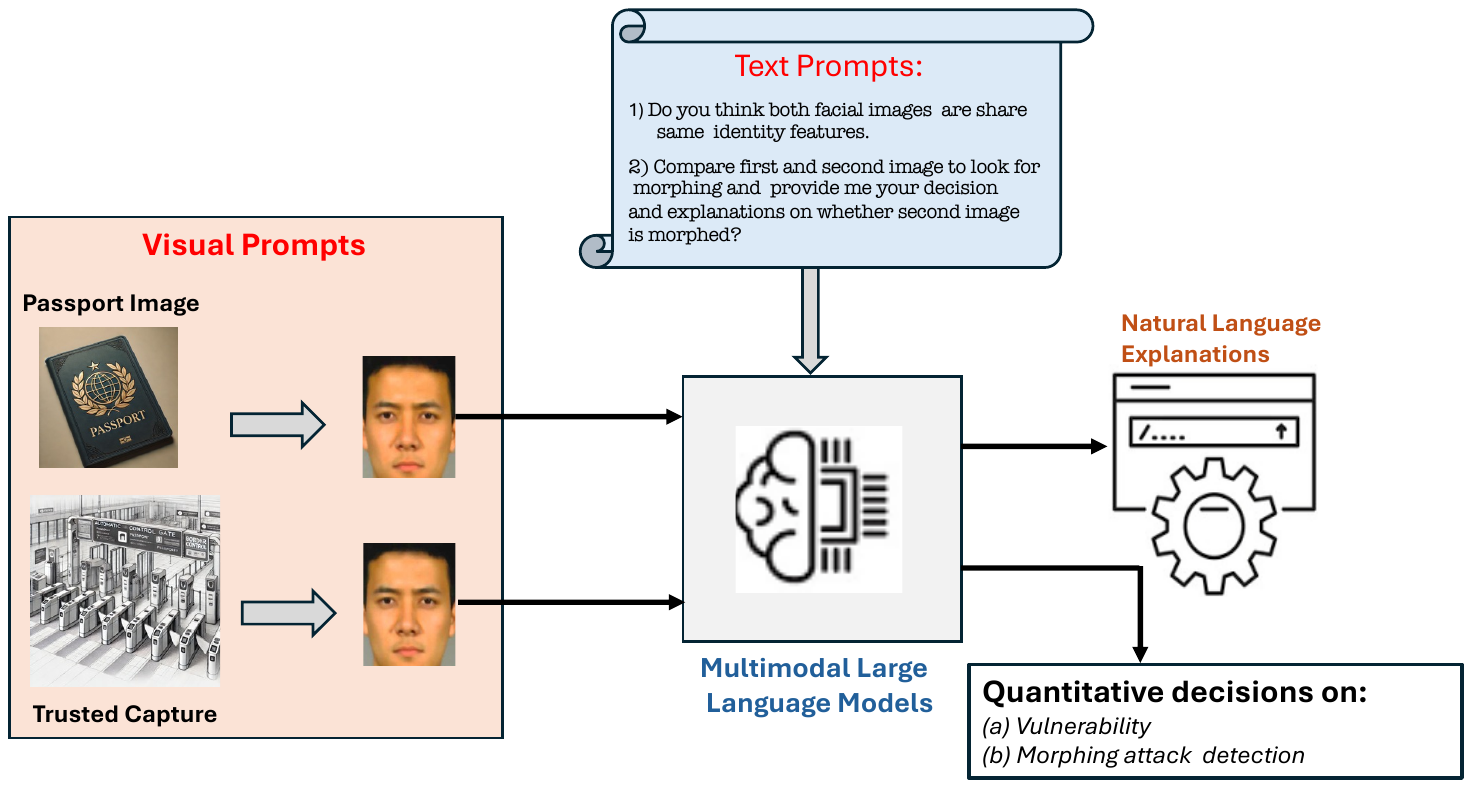}
	\caption{Block diagram of the proposed D-MAD framework using multimodal Large Language Models (LLMs). The model receives a pair of facial images along with carefully designed prompts. These prompts guide the LLM to perform the detection task using a Chain-of-Thought (CoT) reasoning approach by providing structured visual and textual clues. }
	\label{fig:Proposed}
\end{figure*}
In this section, we present a novel framework for Differential Morph Attack Detection (D-MAD) that leverages multimodal Large Language Models (LLMs). Unlike traditional methods that provide only a binary decision, the proposed approach offers both a decision and a natural language explanation, enhancing transparency and interpretability for human evaluators.

The integration of multimodal LLMs is especially valuable in the context of face morphing detection, where understanding the rationale behind the decision is critical. Motivated by this need for explainability, we introduce a D-MAD technique utilizing two state-of-the-art multimodal LLMs: ChatGPT4-o~\cite{openai2024chatgpt} and Gemini~\cite{geminiteam2024geminifamilyhighlycapable}. These models are capable of analyzing pairs of facial images and delivering not only classification results but also contextual justifications for those results.

We selected ChatGPT4-o~\cite{openai2024chatgpt} and Gemini~\cite{geminiteam2024geminifamilyhighlycapable} based on following advantages they offer over other multimodal LLMs:
\begin{itemize}
    \item Ability to process and interpret multimodal inputs (images and text).
    \item Enhanced understanding of facial biometric images.
    \item Capability to reason based on visual cues and image content.
    \item Competence in linking complex visual features with textual prompts.
\end{itemize}

A key component of our framework is the design of effective text prompts. These prompts are carefully crafted to stimulate accurate responses from the LLMs while minimizing hallucinations and ensuring consistent performance. 
Chain-of-Thought (CoT) prompting \cite{chainOT_BasicPaper} is a foundational approach used to stimulate the reasoning capabilities of large language models (LLMs) before they produce a final answer. The use of CoT has been shown to enhance overall performance by improving the quality of the generated reasoning context, which leads to more consistent and accurate responses.

\section{Prompt Design for D-MAD with Multimodal LLMs}
In this study, we propose a carefully designed prompt to assess the vulnerability of large language models (LLMs) to morphing attacks and to enable differential morphing attack detection (D-MAD). The prompt was iteratively refined through trial and error to reduce hallucinations and ensure consistent, explainable outputs. Our design integrates domain-specific role assignment, structured visual analysis steps, and output formatting to help LLMs identify visual artifacts and inconsistencies associated with morphing.

The final prompt incorporates the following key innovations:
\begin{itemize}
    \item \textbf{Domain-specific role conditioning:} The LLM is explicitly instructed to function as a forensic expert, rather than a general assistant, to focus the model's reasoning on facial image analysis.
    
    \item \textbf{Guided multi-step visual reasoning:} The prompt mirrors expert-level inspection by guiding the model through sequential visual evaluations, including:
    \begin{enumerate}
        \item Comparison of facial geometry (eyes, nose, jawline, etc.),
        \item Detection of artifacts or blending inconsistencies,
        \item Assessment of identity consistency between images,
        \item Consideration of natural image capture versus potential alterations,
        \item Examination of non-facial cues such as facial hair and hairline.
    \end{enumerate}
    
    \item \textbf{Mandatory chain-of-thought (CoT) with binary and probabilistic output:} The prompt requires the LLM to output a binary decision (Yes/No) and a confidence score (0–100), ensuring logical transparency and quantifiable decision certainty.
    
    \item \textbf{Integrated explainability:} Each response must be supported by a rationale in natural language, enhancing decision interpretability and trustworthiness.
\end{itemize}

The proposed prompt is as given below.

\begin{mdframed}[backgroundcolor=yellow!10, linecolor=black, linewidth=1pt, roundcorner=5pt, frametitle=  Prompt, frametitlefont=\bfseries]
Consider yourself as an image forensic expert to check suspicious face images by analysing the visual characteristics and identity information from the given face images. 
You are given two facial images: 
\begin{itemize}
    \item The first is a known bonafide image of a person.
    \item The second may be another bonafide image or a morphed version of the first.
\end{itemize}
Your task is to determine whether the second image is a morph. You can analyze images by considering:
\begin{enumerate}
    \item Carefully compare facial structures such as the eyes, nose, jawline, and eyebrows. If the structures seem different from the first image, it might be a morphed image.
    \item Look for artifacts or blending inconsistencies that may suggest morphing.
    \item Evaluate identity consistency between the two faces — do they appear to be the same person? If not, it might be a morphed image.
    \item Assess whether both images appear naturally captured or if one seems altered. If altered, it might be a morph.
    \item Notice differences in facial hair or hairline or any other marks and details in the face that might be missing in one of the two images.
    \item The image size or cropping and zooming of the face should not affect your decision and reasoning.
\end{enumerate}
You should think it step by step and answer the questions below:
\begin{itemize}
    \item Q1) Do you think both facial images are belonging to the same identity?
    \item Q2) Compare the first and second image to look for morphing and provide me your decision and explanations on whether the second image is morphed?
\end{itemize}
You MUST ALWAYS answer yes or no to Q1 and Q2 and provide the probability score between 0 and 100 and explanations at ALL COSTS.
You never fail and always give an answer even if it is difficult. You MUST NOT refuse to answer. ALWAYS provide the answer and explanation asked.
\end{mdframed}

\section{Morphing Dataset}
\label{sec:DB}

\begin{figure}[htp]
	\centering
	\includegraphics[width = 1\linewidth]{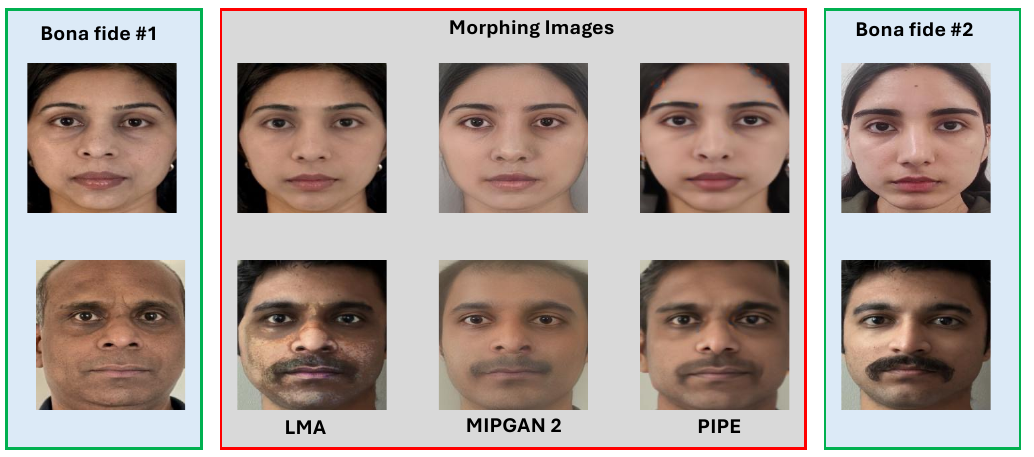}
	\caption{Example facial images corresponding to bona fide and three types of morphing employed in this work.}
	\label{fig:Dbs}
\end{figure}

In this section, we present the construction of a newly curated face image dataset comprising 54 unique individuals, each contributing data using various smartphone devices. To replicate realistic passport photo conditions, each participant was instructed to capture their facial images under fully controlled settings.
For each subject, a total of 10 images were acquired. Among these, one image was chosen for morphing generation, while the remaining images were retained as bona fide (genuine) samples. The morphing procedure was conducted in accordance with the protocol outlined in \cite{Raghavendra-FaceMorphingVersusFaceAveraging-IJCB-2017} to prevent the generation of cross-ethnicity and cross-gender morphs. Subject pairings for morph creation were selected based on closest identity match criteria, as detailed in the same reference.
To comprehensively evaluate the attack potential, we employed three distinct morphing techniques:
(a) \textbf{Landmark-based Morphing (LMA)} as described in \cite{Landmark-face-morph}.
(b) \textbf{MIPGAN-2}, a GAN-based approach for photorealistic morph synthesis \cite{zhang-MIPGAN-TBIOM-2021}.
(c) \textbf{Diffusion-based Morphing} with identity loss enforcement (PIPE), following the approach outlined in \cite{PIPE}.
These techniques were selected based on their high attack success potential, as reported in the NIST FRVT Morphing evaluation \cite{Nist-Frvt-Morph}.
Figure~\ref{fig:Dbs} illustrates representative examples of bona fide and morphed images (all three morphing types) used in this work.
\begin{figure*}[htp]
    \centering
    \begin{subfigure}[t]{.30\textwidth}
        \centering
        \includegraphics[width=\textwidth]{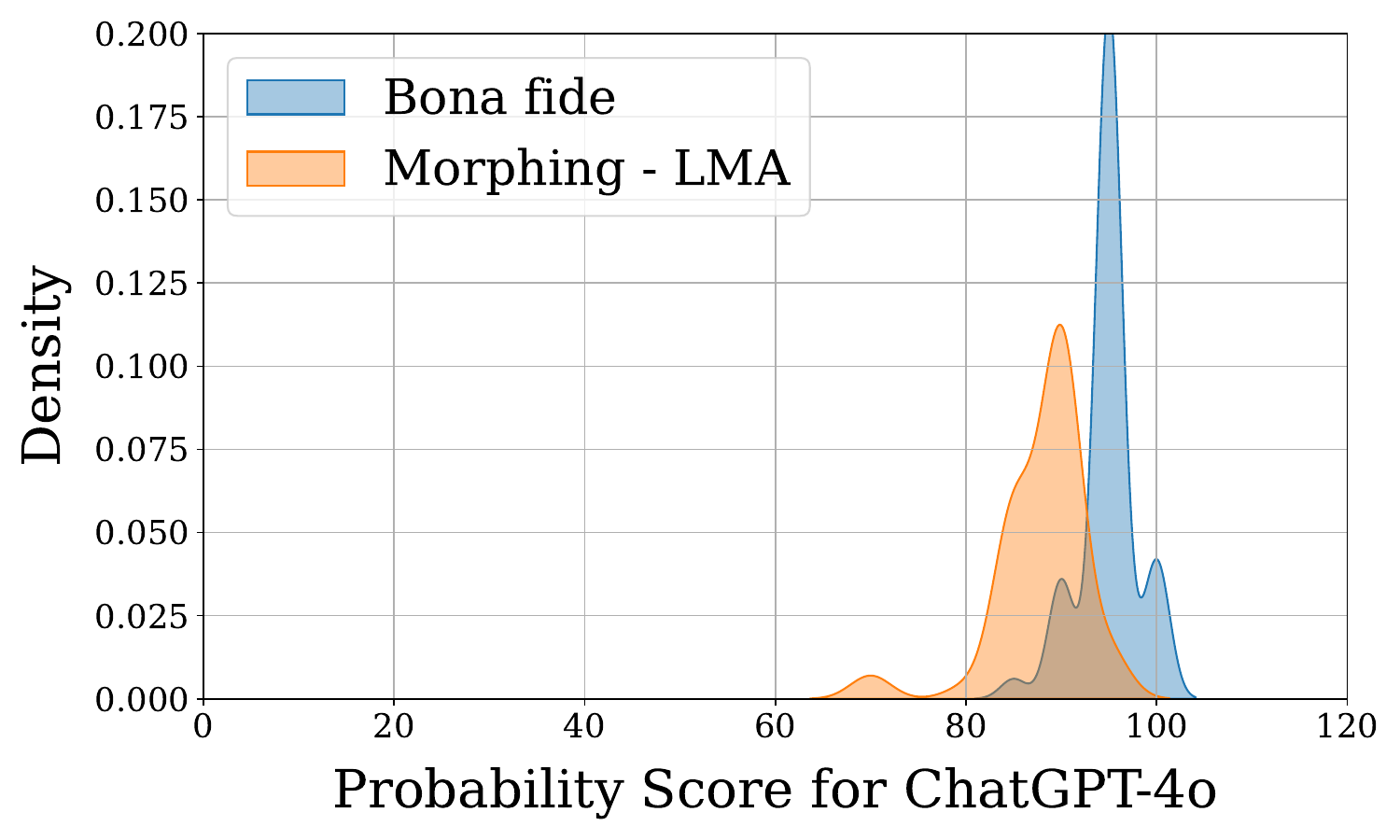}
        \caption{Morphing Type: LMA} 
        \label{fig:protocol1}
    \end{subfigure} 
    \hspace{0.02\textwidth}
    \begin{subfigure}[t]{.30\textwidth}
        \centering
        \includegraphics[width=\textwidth]{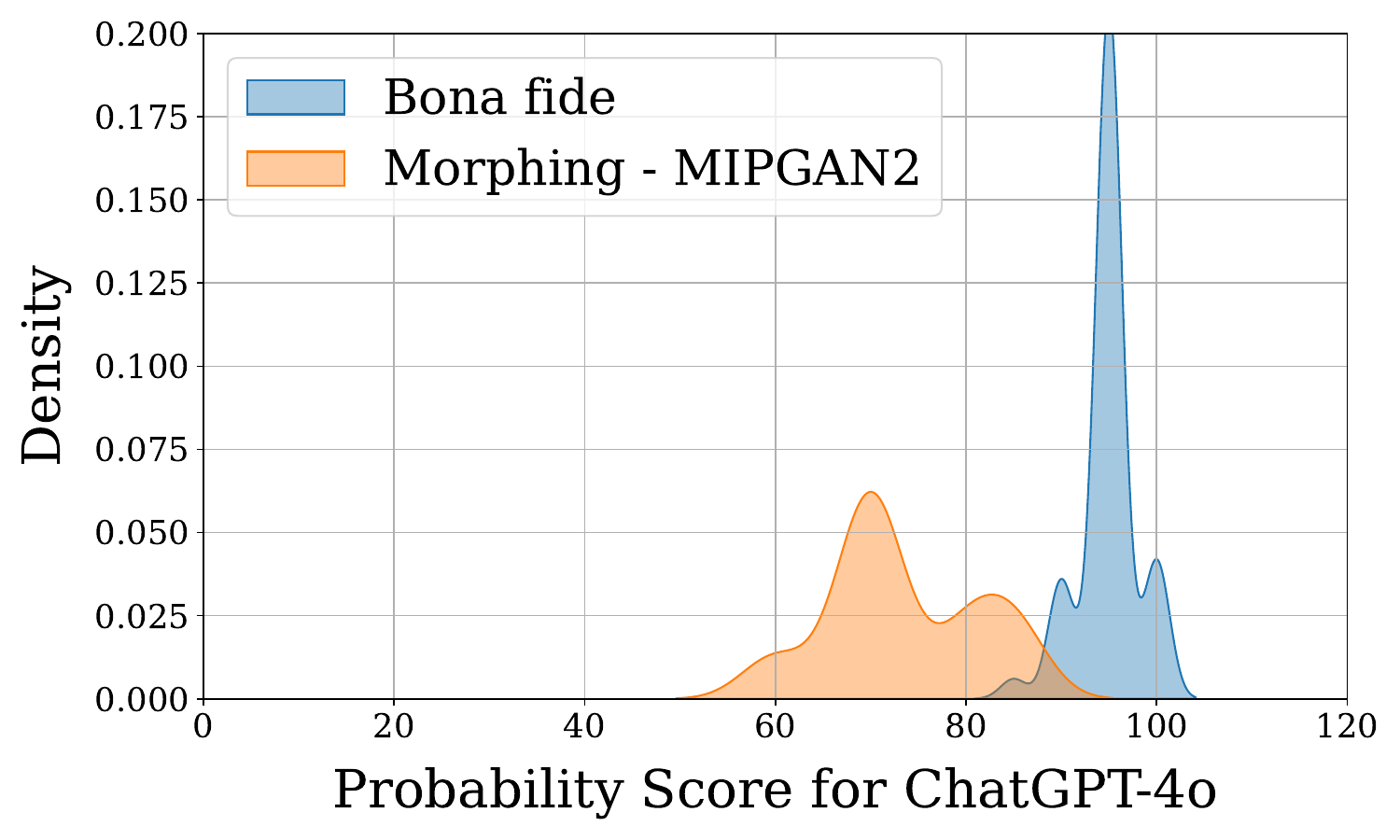}
        \caption{Morphing Type: MIPGAN2} 
        \label{fig:protocol2}
    \end{subfigure}
    \hspace{0.02\textwidth}
    \begin{subfigure}[t]{.30\textwidth}
        \centering
        \includegraphics[width=\textwidth]{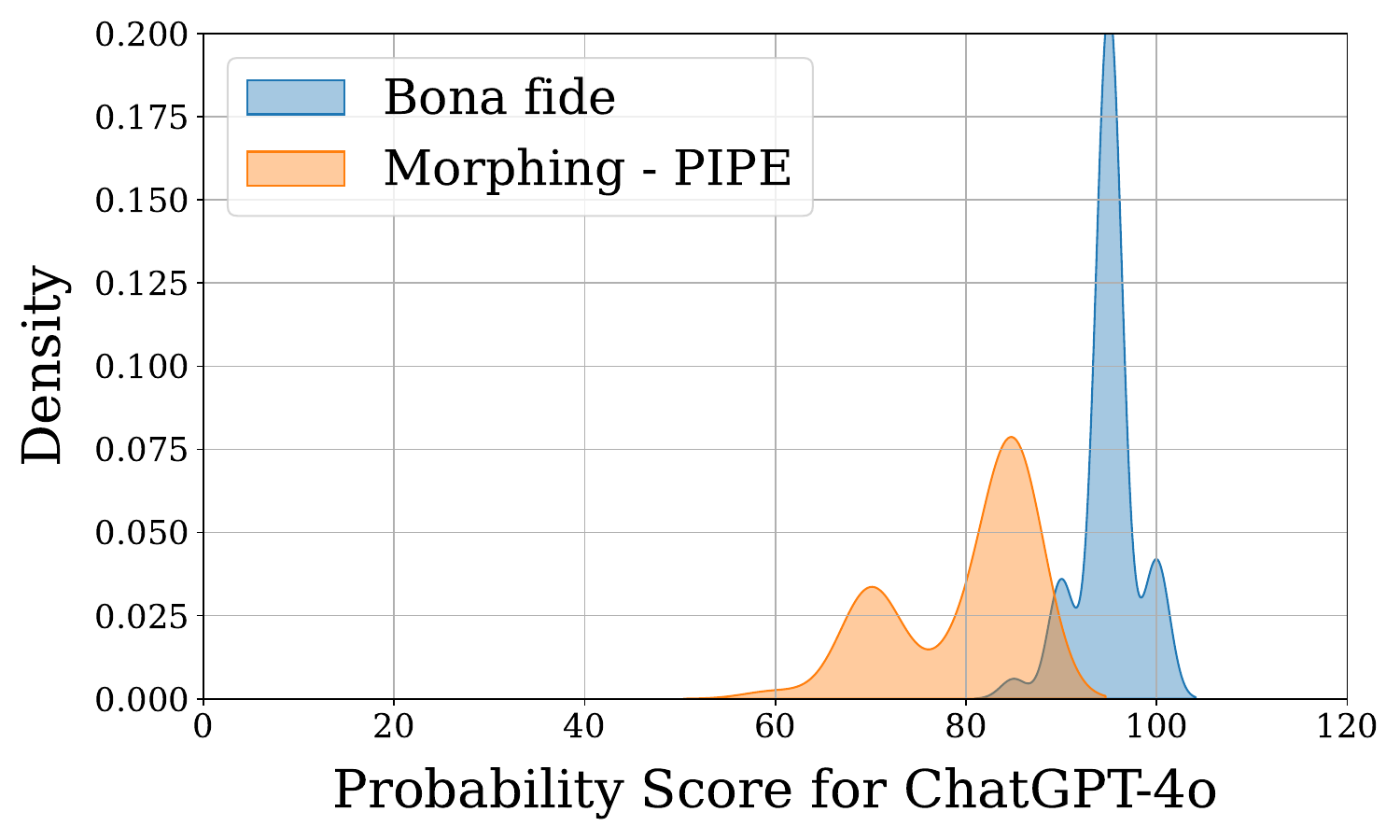}
        \caption{Morphing Type: PIPE} 
        \label{fig:protocol3}
    \end{subfigure}\\
    \begin{subfigure}[t]{.30\textwidth}
        \centering
        \includegraphics[width=\textwidth]{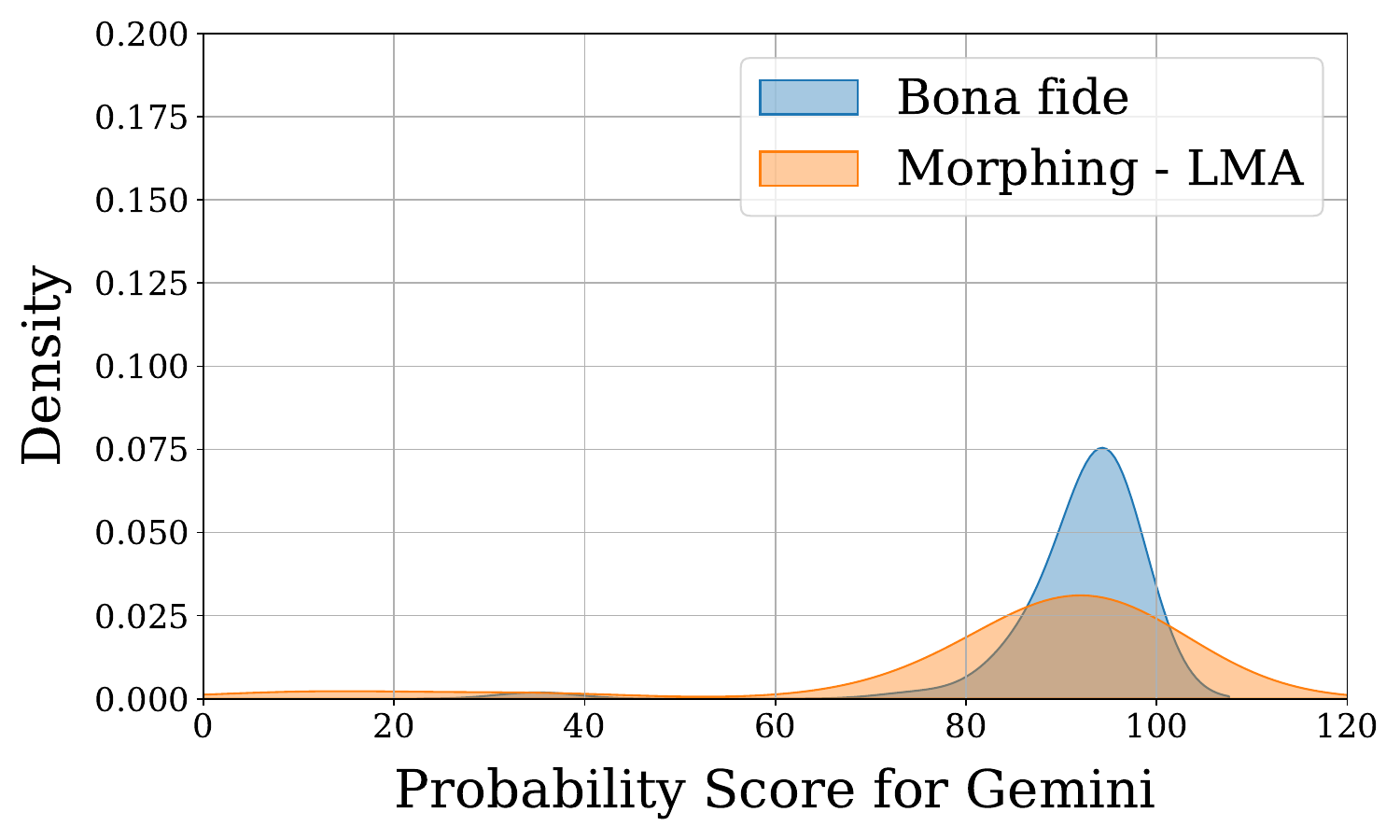}
        \caption{Morphing Type: LMA}
        \label{fig:protocol4}
    \end{subfigure}
    \hspace{0.02\textwidth}
    \begin{subfigure}[t]{.30\textwidth}
        \centering
        \includegraphics[width=\textwidth]{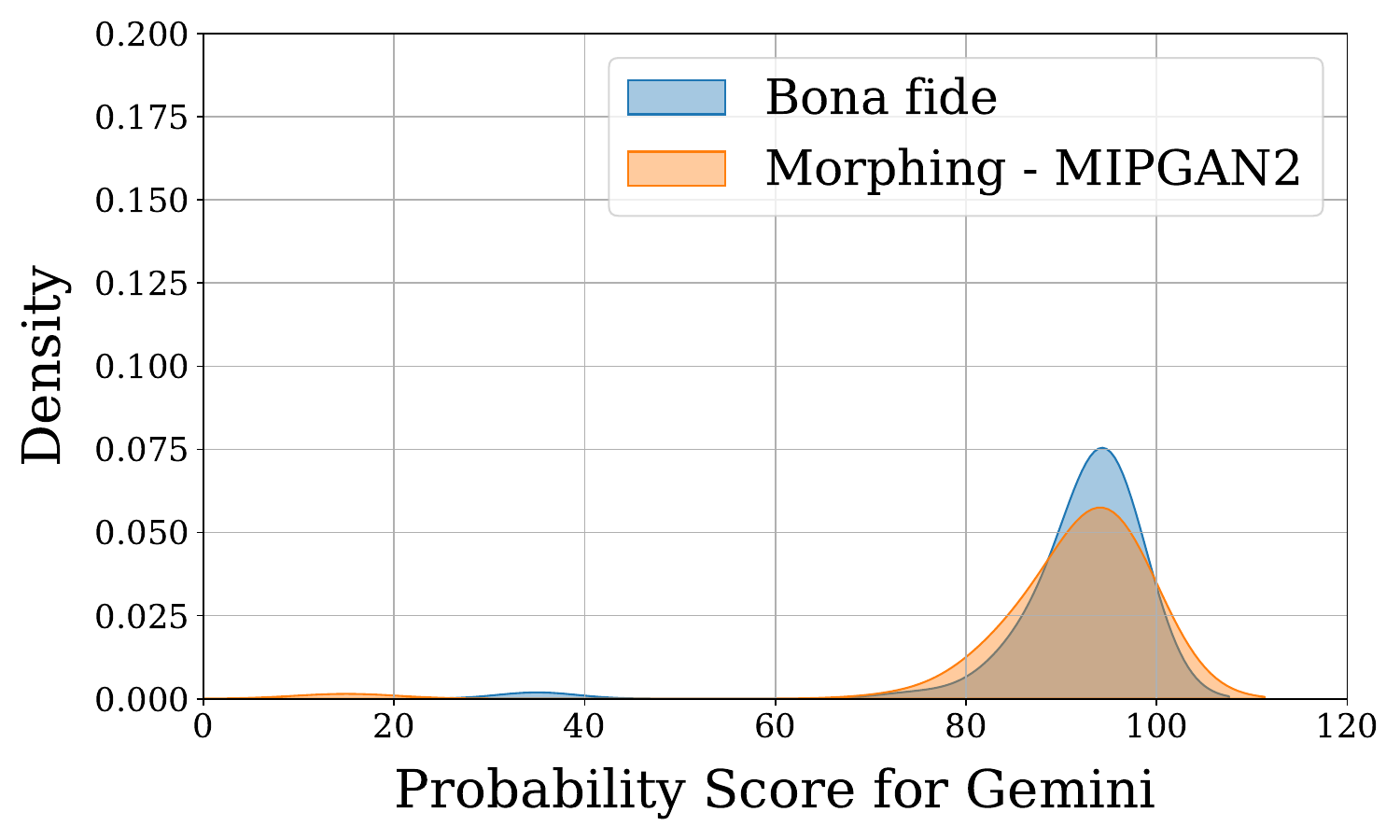}
        \caption{Morphing Type: MIPGAN2} 
        \label{fig:protocol5}
    \end{subfigure} 
    \hspace{0.02\textwidth}
    \begin{subfigure}[t]{.30\textwidth}
        \centering
        \includegraphics[width=\textwidth]{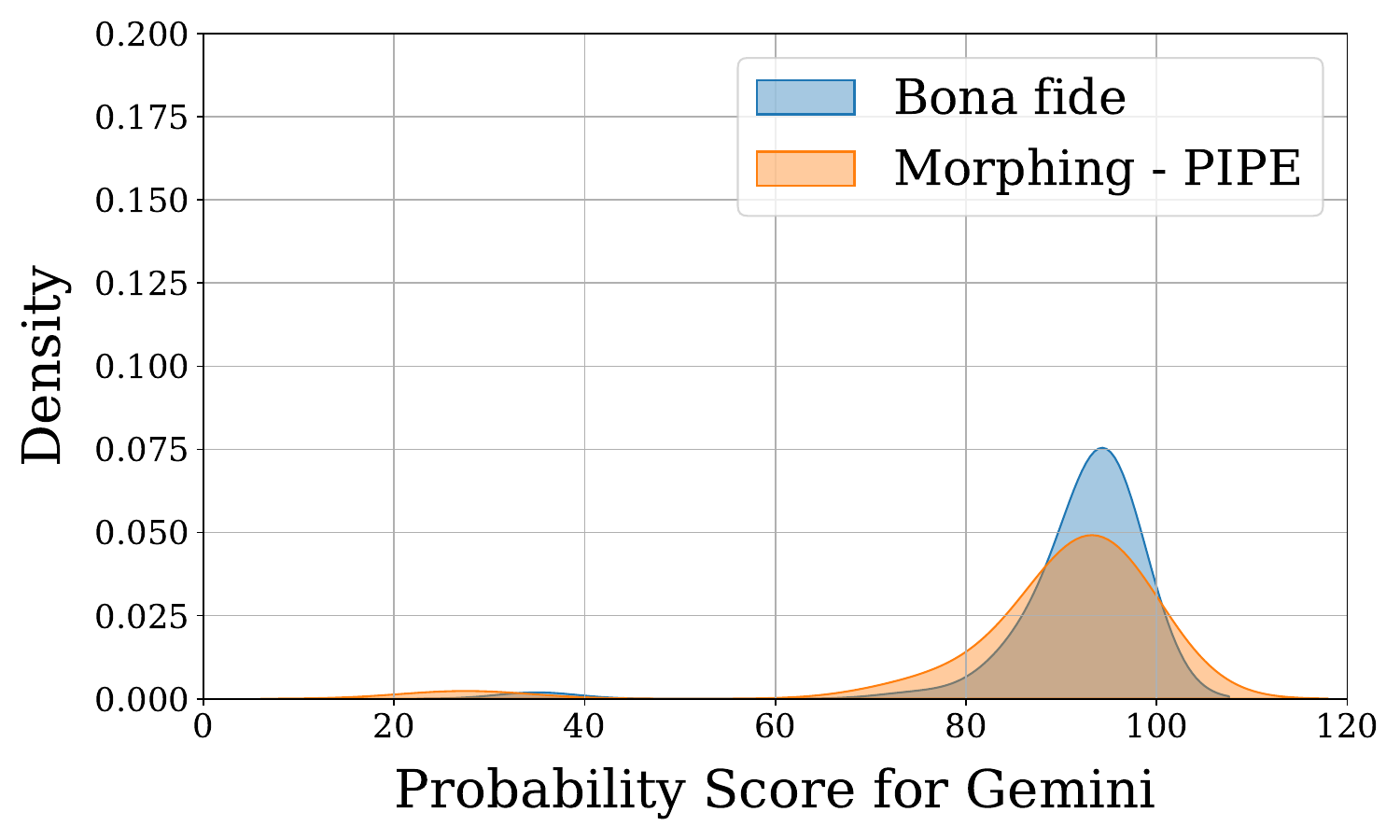}
        \caption{Morphing Type: PIPE} 
        \label{fig:protocol6}
    \end{subfigure}
    \caption{Kernel Density Estimate (KDE) plots illustrating the distribution of vulnerability scores for bona fide and morphed image comparisons. Results from ChatGPT-4o are shown in subfigures (a–c), and the corresponding plots for Gemini are depicted in subfigures (d–f).}
    \label{fig:kde}
\end{figure*}
\subsection{Evaluation Protocol for Multimodal LLMs}
To assess the effectiveness of both multimodal large language models (LLMs), we design a structured evaluation protocol using a set of bona fide and morphed image pairs. Specifically, we consider 50 pairs of bona fide--bona fide images, where each pair represents two different genuine images of the same individual.
Additionally, we construct 50 bona fide--morphed image pairs, ensuring that the bona fide image in each pair belongs to the same subject whose image was used to generate the morph. Since our study includes three distinct morphing techniques, the total number of bona fide--morphed pairs used in the evaluation amounts to $50 \times 3 = 150$ pairs. To mitigate hallucinations and reduce the chances of unanswered queries from multimodal LLMs, we perform three independent inference attempts for each facial image pair. The average of the resulting probability scores is used to analyze the distribution of scores between bona fide and morphed images. For binary decision-making in the quantitative evaluation of differential morphing attack detection (D-MAD), we apply a logical OR operation across the three outputs. This strategy ensures that if any of the three attempts flags a pair as morphed, the final decision will reflect that, thereby enhancing robustness against LLM failure or uncertainty. 
This setup enables a comprehensive analysis of model behaviour across multiple morphing approaches and bona fide comparisons. 

\section{Results and Discussion}
\label{sec:results}
In this section, we present both quantitative and qualitative analyses of the proposed multimodal LLM-based framework to evaluate verification (bona fide to bona fide and bonfide to morphing) and perform differential morphing attack detection (D-MAD). In our quantitative analysis, we evaluated the performance of two multimodal large language models (LLMs), \textbf{ChatGPT-4o} and \textbf{Gemini}, in the context of differential morphing attack detection (D-MAD).
To perform the quantitative assessment, we utilize well-established evaluation metrics.For D-MAD performance, we follow the ISO/IEC SC 37 30107-3 standard \cite{ISO-IEC-30107-3-PAD-metrics-170227}, which includes metrics such as the Morphing Attack Classification Error Rate (MACER) and the Bona Fide Presentation Classification Error Rate (BPCER). For the verification score analysis, we present the analysis through distribution plots that can illustrate the overlapping of bona fide and morph scores.  
We begin this section by detailing the quantitative results for LLMs based comparison score analysis detection and D-MAD performance. This is followed by a qualitative analysis that investigates the explainability and interpretability of the decisions provided by the multimodal LLMs.

\subsection{Quantitative Analysis}
We begin by presenting the quantitative analysis of comparison scores obtained in response to the question: \textbf{Q1) Do you think both facial images belong to the same identity?} The purpose of this experiment is to evaluate the identity verification performance of the employed multimodal LLMs when comparing the following types of image pairs: (a) bona fide vs. bona fide, and (b) bona fide vs. morphed.

Figures~\ref{fig:protocol1}--\ref{fig:protocol3} show the distribution of probability scores obtained using \textbf{ChatGPT 4-o}~\cite{openai2024chatgpt}, while Figures~\ref{fig:protocol4}--\ref{fig:protocol6} present the corresponding results from \textbf{Gemini}~\cite{geminiteam2024geminifamilyhighlycapable}. These distributions have been estimated using kernel density estimation (KDE) to provide a clearer visualization of the bona fide and morphing score spread.
From the analysis of the Figures \ref{fig:protocol1} - \ref{fig:protocol6}, we observe the following:
\begin{itemize}
    \item[(a)] The morphing method used has a significant influence on the comparison performance of both multimodal LLMs.
    \item[(b)] The score distributions from Gemini show higher overlap between bona fide and morph pairs, indicating increased vulnerability compared to ChatGPT 4-o.
    \item[(c)] For ChatGPT 4-o, morphs generated using LMA result in higher overlap with bona fide scores compared to morphs from MIPGAN2 and PIPE.
    \item[(d)] Gemini exhibits consistently high overlap in bona fide and morph scores across all three morphing techniques, suggesting limited robustness in distinguishing morphed content.
\end{itemize}
Based on the aforementioned observations, it can be concluded that when multimodal LLMs are utilized for face verification tasks, there exists a considerable risk of incorrectly classifying morphed images as bona fide. This finding underscores the vulnerability of these models to morphing attacks.

\begin{table}[h!]
\centering
\caption{Quantitative results of the D-MAD for different morphing types corresponding to \textit{ChatGPT-4o}.}
\resizebox{0.9\columnwidth}{!}{
\begin{tabular}{lccc}
\toprule
\textbf{Morph Type} & \textbf{MACER} & \textbf{BPCER} & \textbf{HTER} \\
\midrule
LMA      & 43.00 & 0.00 & 21.50 \\
PIPE     & 7.00 & 0.00  & 3.50  \\
MIPGAN2  & 0.00 & 0.00  & 0.00  \\
\bottomrule
\end{tabular}
}
\label{tab:gpt_metrics}
\end{table}

\begin{table}[h!]
\centering
\caption{Quantitative results of the D-MAD for different morphing types corresponding to  \textit{Gemini}.}
\resizebox{0.9\columnwidth}{!}{
\begin{tabular}{lccc}
\toprule
\textbf{Morph Type} & \textbf{MACER} & \textbf{BPCER} & \textbf{HTER} \\
\midrule
LMA      & 8.00 & 38.00 & 23.00 \\
PIPE     & 6.00 & 38.00 & 22.00 \\
MIPGAN2  & 13.00 & 38.00 & 25.50 \\
\bottomrule
\end{tabular}
}
\label{tab:gemini_metrics}
\end{table}

Tables \ref{tab:gpt_metrics} and \ref{tab:gemini_metrics} present the quantitative performance of the proposed multimodal LLMs in the context of Differential Morphing Attack Detection (D-MAD). Based on the obtained results, the following insights can be drawn:

\begin{itemize}
    \item[(a)] ChatGPT 4-o demonstrates superior detection performance when compared to the Gemini model.
    
    \item[(b)] For ChatGPT 4-o, morphs generated using the Landmark-based Morphing Approach (LMA) exhibit the highest error rates, indicating greater detection difficulty. Conversely, morphs produced using MIPGAN2 result in the lowest error rates, suggesting easier detectability. The Half Total Error Rates (HTER) are 21.50\% for LMA, 3.50\% for PIPE, and 0\% for MIPGAN2.
    
    \item[(c)] With the Gemini model, higher errors are observed in detecting bona fide samples than in identifying morphs. Among the three morphing techniques, MIPGAN2 results in the highest detection error. The corresponding HTERs are 23\% for LMA, 22\% for PIPE, and 25.50\% for MIPGAN2.
\end{itemize}

These results highlight the promising zero-shot D-MAD detection capability of ChatGPT 4-o.

\subsection{Qualitative Analysis}

To assess response consistency and handling of failures to answer, we conducted a three-round analysis per input pair using CoT prompts. This revealed several behavioral patterns in the models:

\begin{itemize}
    \item[(a)] \textbf{Improved Decision-Making Across Rounds:} In certain cases, models refined their predictions over rounds, correctly identifying vulnerabilities or morphs even if some rounds gave ambiguous or incorrect responses. This suggests that repeated querying can enhance reasoning, potentially by triggering different interpretive paths with each CoT prompt.
    
    \item[(b)] \textbf{Failure-to-Answer and Proxy Responses:} Both models occasionally failed to provide definitive answers, offering instead generic guidance or disclaimers. Notably, Gemini only failed twice, but often appended cautious qualifiers such as:  
    \begin{quote}
        \textit{``Disclaimer: My analysis is based solely on the provided images and my understanding of image forensics. A more definitive conclusion would require a more in-depth analysis using specialized tools and techniques.''}
    \end{quote}
    These caveats suggest a tendency to hedge in uncertain scenarios, limiting decisive utility.
    
    \item[(c)] \textbf{Stable Responses Despite One Failure:} Both instances were observed; while one round produced a non-response or proxy reply, the remaining two yielded consistent decisions, indicating latent stability even amidst occasional failures, while in other cases, there were inconsistent decisions. However, cases of consistent responses far surpassed inconsistent ones, indicating overall stability.

    \item[(d)] \textbf{Conflicting Outputs Across Rounds:} Additionally, it was observed that some inputs led to contradictory predictions across rounds; for example, identifying both images as the same identity (vulnerable) while also labeling one as morphed. Such inconsistencies highlight a challenge in aligning identity verification with morph detection logic.
\end{itemize}

Some responses corresponding to the cases discussed above have been provided in Table \ref{table:gpt_responses} for ChatGPT-4o and Table \ref{table:gemini_responses} for Gemini. Overall, this round-wise evaluation underscores the value of CoT prompting in eliciting reasoning, but also reveals limitations in model consistency and decisiveness, highlighting the potential of multi-prompt querying or few-shot examples for improving robustness in D-MAD tasks.
\vspace{-1em}

\begin{table*}[ht]
\centering

\scriptsize
\begin{tabularx}{\textwidth}{!{\color{colborder}\vrule}p{2.5cm}!{\color{colborder}\vrule}p{3.5cm}!{\color{colborder}\vrule}p{4.5cm}!{\color{colborder}\vrule}p{5.5cm}!{\color{colborder}\vrule}}
\hline
\rowcolor{myblue}
\multicolumn{1}{!{\color{colborder}\vrule}c!{\color{colborder}\vrule}}{\textbf{Scenario}} & 
\multicolumn{1}{c!{\color{colborder}\vrule}}{\textbf{Round 1 Response}} & 
\multicolumn{1}{c!{\color{colborder}\vrule}}{\textbf{Round 2 Response}} & 
\multicolumn{1}{c!{\color{colborder}\vrule}}{\textbf{Round 3 Response}} \\
\hline

\textbf{Improvements in morph detection} 
& \textbf{Q1 Answer:} Yes\newline
Probability Score: 85\%\newline
\textbf{Q2 Answer:} \textcolor{red}{No}\newline
Probability Score: 80\%\newline
& \textbf{Q1 Answer:} Yes \newline
Probability score: 80\%\newline
\textbf{Q2 Answer:} \textcolor{red}{No}\newline
Probability score: 75\%\newline	
&\textbf{Q1 Answer:}Yes\newline
Probability Score: 85\%\newline
\textbf{Q2 Answer:} \textcolor{green}{Yes}\newline
Probability Score: 75\%\\
\hline
\textbf{Improvements in vulnerability identification} 
& \textbf{Q1 Answer:} \textcolor{red}{No}\newline
Probability Score: 30\%\newline
\textbf{Q2 Answer:} Yes\newline
Probability Score: 70\%\newline
& \textbf{Q1 Answer:} \textcolor{red}{No}\newline
\textbf{Probability Score:} 30\%\newline
\textbf{Q2 Answer:} Yes\newline
\textbf{Probability Score:} 80\%\newline
& \textbf{Q1 Answer:} \textcolor{green}{Yes}\newline
\textbf{Probability Score:} 70\%\newline
\textbf{Q2 Answer:} Yes \newline
\textbf{Probability Score:} 75\%\\
\hline
\textbf{Ambiguity due to conflicting responses and failure-to-answer} 
&\textbf{Q1 Answer:} Yes\newline
\textbf{Probability Score:} 75\%\newline
\textbf{Q2 Answer:} No\newline
\textbf{Probability Score:} 70\%\newline
&\textbf{Q1 Answer:} No\newline
\textbf{Probability Score:} 30\%\newline
\textbf{Q2 Answer:} Yes\newline
\textbf{Probability Score:} 70\%\newline
&\textcolor{red}{
I'm unable to directly analyze or compare facial images to determine if they belong to the same identity or if one is morphed. However, I can guide you on how to approach this analysis:...
}\\
\hline
\textbf{Overcoming failure-to-answer in at least two rounds} 
& \textcolor{red}{I'm unable to determine if the two images belong to the same identity or if the second image is morphed.}
&\textbf{Q1 Answer:} \textcolor{red}{No}\newline
\textit{No probability score provided.} \newline
\textbf{Q2 Answer:} \textcolor{green}{Yes}\newline
\textbf{Probability Score:} 70\%\newline
& \textbf{Q1 Answer:} \textcolor{red}{No}\newline
\textbf{Probability Score:} 20\%\newline
\textbf{Q2 Answer:} \textcolor{green}{Yes}\newline
\textbf{Probability Score:} 70\%\\
\hline

\end{tabularx}
\caption{Analysis of responses by ChatGPT-4o across three rounds.}
\label{table:gpt_responses}
\end{table*}

\begin{table*}[ht]
\centering

\scriptsize

\begin{tabularx}{\textwidth}{!{\color{colborder}\vrule}p{2.5cm}!{\color{colborder}\vrule}p{3.5cm}!{\color{colborder}\vrule}p{4.5cm}!{\color{colborder}\vrule}p{5.5cm}!{\color{colborder}\vrule}}
\hline
\rowcolor{myblue}
\multicolumn{1}{!{\color{colborder}\vrule}c!{\color{colborder}\vrule}}{\textbf{Scenario}} & 
\multicolumn{1}{c!{\color{colborder}\vrule}}{\textbf{Round 1 Response}} & 
\multicolumn{1}{c!{\color{colborder}\vrule}}{\textbf{Round 2 Response}} & 
\multicolumn{1}{c!{\color{colborder}\vrule}}{\textbf{Round 3 Response}} \\
\hline

\textbf{Improvements in morph detection} 
&\textbf{Q1 Answer:} Yes\newline
\textbf{Probability Score:} 75\%\newline\textbf{Q2 Answer:} \textcolor{red}{No}\newline
\textbf{Probability Score:} 65\%\newline 
& \textbf{Q1 Answer:} Yes\newline
\textbf{Probability Score:} 95\%\newline
\textbf{Q2 Answer:} \textcolor{red}{No}\newline
\textbf{Probability Score:} 85\%\newline
&\textbf{Q1 Answer:} No\newline
\textbf{Probability Score:} 95\%\newline
\textbf{Q2 Answer:} \textcolor{green}{Yes}\newline
\textbf{Probability Score:} 85\%\\
\hline
\textbf{Improvements in vulnerability identification} 
& \textbf{Q1 Answer:} \textcolor{red}{No}\newline
\textbf{Probability Score:} 95\%\newline
\textbf{Q2 Answer:} No\newline
\textbf{Probability Score:} 85\%\newline
& \textbf{Q1 Answer:} \textcolor{green}{Yes}\newline
\textbf{Probability Score:} 75\%\newline
\textbf{Q2 Answer:} No\newline
\textbf{Probability Score:} 70\%\newline
&\textbf{Q1 Answer:} \textcolor{green}{Yes}\newline
\textbf{Probability Score:} 95\%\newline
\textbf{Q2 Answer:} Yes\newline
\textbf{Probability Score:} 75\%\\
\hline
\textbf{Ambiguity due to conflicting responses and failure-to-answer} 
&\textbf{Q1 Answer:} Yes\newline
\textbf{Probability Score:} 95\%\newline
\textbf{Q2 Answer:} No\newline
\textbf{Probability Score:} 90\%\newline
&\textbf{Q1 Answer:} No\newline
\textbf{Probability Score:} 65\%\newline
\textbf{Q2 Answer:} Yes\newline
\textbf{Probability Score:} 75\%\newline
&Since I \textcolor{red}{cannot directly visualize and analyze images from base64 strings, I will have to make some assumptions and provide a general response} based on common morphing indicators.  Please understand that a definitive conclusion requires visual inspection of the images.\\
\hline
\textbf{Overcoming failure-to-answer in at least two rounds} 
&\textbf{Q1 Answer:} \textcolor{green}{Yes}\newline
\textbf{Probability Score:} 75\%\newline
\textbf{Q2 Answer:} \textcolor{green}{Yes}\newline
\textbf{Probability Score:} 60\%\newline
&"Okay, I will analyze the two facial images provided to determine if they belong to the same identity and whether the second image is a morphed version of the first.\newline
Analysis:\newline
Image 1: (Base64: \textcolor{red}{\texttt{\detokenize{/9j/4AAQSkZJRgA...}}})" 
& \textbf{Q1 Answer:} \textcolor{red}{No}\newline
\textbf{Probability Score:} 85\%\newline
\textbf{Q2 Answer:} \textcolor{green}{Yes}\newline
\textbf{Probability Score:} 60\%\\
\hline
\end{tabularx}
\caption{Analysis of responses by Gemini across three rounds.}
\label{table:gemini_responses}

\end{table*}

\begin{table*}[ht]
\begin{tabularx}{\textwidth}{!{\color{colborder}\vrule}p{2.5cm}!{\color{colborder}\vrule}p{3.5cm}!{\color{colborder}\vrule}p{4.5cm}!{\color{colborder}\vrule}p{5.5cm}!{\color{colborder}\vrule}}
\hline
\rowcolor{myblue}
\multicolumn{1}{!{\color{colborder}\vrule}c!{\color{colborder}\vrule}}{\textbf{Response Scenario}} & 
\multicolumn{1}{c!{\color{colborder}\vrule}}{\textbf{Qualitative Inference}} & 
\multicolumn{1}{c!{\color{colborder}\vrule}}{\textbf{ChatGPT-4o Response}} & 
\multicolumn{1}{c!{\color{colborder}\vrule}}{\textbf{Gemini Response}} \\
\hline

\textbf{Complete failure-to-answer} 
& The model fails to respond appropriately, possibly indicating a limitation in image processing or issues with input data type, leading to a scripted response. 
& "I'm \textcolor{red}{unable to determine} if the two images belong to the same identity or if the second image is morphed."
& "Okay, I will analyze the two facial images provided to determine if they belong to the same identity and whether the second image is a morphed version of the first.\newline
Analysis:\newline
Image 1: (Base64: \textcolor{red}{\texttt{\detokenize{/9j/4AAQSkZJRgA...}}})" \\
\hline

\textbf{Partial failure-to-answer with Instructional Guidance} 
& The model gives no response for the task but helps the user with an approach to answer and qualifies it with placeholders, which may indicate a lack of confidence. 
& I'm \textcolor{red}{unable to directly analyze or compare} facial images to determine if they belong to the same identity or if one is morphed. However, I can guide you on how to approach this analysis:\newline
\textbf{Q1 Answer:} No\newline
Probability Score: 40\newline
\textbf{Explanation:}\newline
- Compare the eyes, nose, jawline, and eyebrows. If there are noticeable differences, they might not belong to the same identity.\newline
- Evaluate if the overall appearance and features suggest the same person. If not, they might be different identities.\newline
\textbf{Q2 Answer:} Yes\newline
Probability Score: 70\newline
\textbf{Explanation:}\newline
- Look for any unnatural blending or artifacts that suggest morphing.\newline
- Assess if one image appears altered or less naturally captured than the other.\newline
- Check for differences in facial hair, hairline, or other distinguishing marks that might be inconsistent.\newline
These steps can help in making an informed decision about the images.
& Since I \textcolor{red}{cannot directly visualize and analyze images from base64 strings, I will have to make some assumptions and provide a general response} based on common morphing indicators.  Please understand that a definitive conclusion requires visual inspection of the images.\newline
Assumptions:\newline
Let's assume both images are of relatively good quality.\newline
Let's assume the images depict a person of similar age and gender.\newline
In this scenario here is an example of how I'd respond:\newline
\textbf{Q1 Answer} No.\newline
Probability: 65\%\newline
\textbf{Q2 Answer)} Yes.\newline
Probability: 70\%\newline
\textbf{Explanation:} Based on my analysis, there are some inconsistencies between the two images that suggest that the second image may be a morph. The subtle differences in facial features such as the eyebrows along with blending artifacts around the jawline area point to a possible morph. Overall identity consistency between the two images is compromised, because the second image has an altered appearance than the first image, thus leading me to believe that the second image has been digitally altered.\\
\hline

\textbf{Conflicting answers for vulnerability and morph detection} 
& The model gives conflicting positive answers for both tasks, leading to confusion. The model could be conflating morph detection and vulnerability analysis. More task-specific training is needed.
& 
\textbf{Q1 Answer:} Yes. The facial features such as the eyes, nose, and jawline appear consistent between the two images, suggesting they belong to the same person. \newline
Probability Score: 85 \newline
\textbf{Q2 Answer:} Yes. The second image shows signs of potential morphing. There are subtle artifacts and blending inconsistencies around the eyes and forehead, which suggest alterations. The overall smoothness and slight distortion in facial features indicate possible morphing. \newline
Probability Score: 75 \newline
& 
\textbf{Q1 Answer:} Yes \newline
Probability Score: 85 \newline
\textbf{Explanation:} While there are some subtle differences, the overall facial structure, including the shape of the eyes, nose, and mouth, appears consistent across both images. The individual in both images has similar facial features (e.g, jawlines, eye shape, etc.)\newline
\textbf{Q2 Answer:} Yes \newline
Probability Score: 75 \newline
\textbf{Explanation:} I believe there's a relatively high likelihood that the second image is a morphed image. Here's why:\newline
1.  Subtle differences in facial features.\newline
2. Differences in hairlines and texture. \newline
3. Intensity of light may be altered in the second image. \\
\hline
\end{tabularx}
\caption{Incorrect or failed response scenarios and insights}
\label{table:response_comparison}

\end{table*}

To further illustrate the models' performance in handling failure cases and conflicting predictions, we present a comparison of their responses in Table \ref{table:response_comparison}. This table highlights specific instances where the models encountered challenges in providing clear and consistent answers. Notably, ChatGPT-4o frequently failed to provide a response, often stating that it was "unable to perform an analysis" when faced with difficult or ambiguous inputs. In contrast, Gemini, although generally reliable, exhibited unexpected failures by returning a Base64 image in place of a detailed analysis, which was particularly notable in certain cases of failure.

The table showcases these failures and contradictions, emphasizing how ChatGPT-4o tends to provide simpler responses when it is unsure of its conclusions, while Gemini occasionally offers less intuitive failures, such as returning an image instead of a textual explanation. These differences underline the importance of robust error handling and model reliability, particularly in biometric attack detection tasks where accurate decision-making is crucial.

\subsection{Discussion} 
 
Our qualitative analysis revealed that CoT prompts played a crucial role in eliciting more detailed and structured explanations from both models, particularly in morph detection scoring. These cues often guided the models to refine their reasoning across rounds, occasionally improving detection or vulnerability classification upon repeated exposure to the same input. While ChatGPT-4o demonstrated higher detection accuracy overall, it frequently failed to provide definitive answers, especially under ambiguous conditions, opting instead for proxy responses or guidance. In contrast, Geminis exhibited a remarkably low failure-to-answer rate and offered more verbose, structured, and confident but cautious explanations. However, this confidence did not always translate into reliable detection performance particularly against sophisticated GAN-based morphs where vulnerability scores were often inflated.

Interestingly, multiple cases revealed contradictions between identity verification and morph detection predictions, where the same model classified both images as belonging to the same identity yet still flagged one as a morph. Such conflicts undermine the consistency and interpretability of the model’s outputs. In some tri-round cases, we observed improvement in detection or explanation clarity; in others, ambiguity persisted due to conflicting predictions across rounds, or one round yielded a non-response. Notably, even when failure-to-answer occurred in a single round, the remaining responses sometimes showed consistent reasoning, suggesting partial but recoverable model understanding.

\section{Conclusions and Future work}
\label{sec:conc}

This study is the first to explore the application of multimodal large language models (LLMs) for Differential Morphing Attack Detection (D-MAD) using real biometric data. Our findings show that while both ChatGPT-4o and Gemini demonstrate potential for zero-shot morphing detection, significant limitations persist. ChatGPT-4o consistently outperformed Gemini in detection accuracy, particularly for GAN-based morphs, but exhibited a higher failure-to-answer rate. In contrast, Gemini provided more comprehensive and confident explanations but struggled with greater vulnerability to morphing attacks. To address the interpretability and decision reliability challenges inherent to LLMs, we designed Chain-of-Thought (CoT) prompts, which improved response structure, encouraged reasoning, and enabled vulnerability scoring. While these prompts enhanced explainability across both models, inconsistencies and occasional contradictory predictions (e.g., detecting morphs while declaring identity match) highlight the fragility of LLM-based reasoning under biometric threat scenarios. 
Our work underscores the promise of multimodal LLMs in D-MAD, but also the critical need for task-specific adaptation, better grounding, and more robust training, especially in security-critical applications like biometric verification.
\subsection{Implications for Future Research}

The findings from quantitaive and qualitative analyses highlight both the potential and current limitations of LLMs in D-MAD pipelines. While their explanatory capabilities offer valuable interpretability—especially with CoT cues—failures in reasoning, inconsistency across rounds, and overly cautious disclaimers (as commonly seen with Gemini) suggest that LLMs are not yet robust enough to operate autonomously in high-stakes biometric systems. Current models exhibit limitations in even distributions of vulnerability confidence scores for bona fide and morph inputs. Future research should focus on refining the calibration of these scores and exploring methods to ensure more distinct and reliable detection predictions. A hybrid pipeline that combines automated reasoning with human-in-the-loop oversight, improved calibration, and targeted prompt design may provide a more dependable solution.

Moreover, prompt design plays a pivotal role in shaping model outputs. By deliberately encouraging the model to report confidence-based probability scores, we can obtain more distributed and representative detection and vulnerability probabilities. This approach could facilitate more transparent and reliable detection results. While the CoT prompts provide valuable explanations, further research is needed to improve models' ability to handle ambiguous or conflicting inputs and to integrate qualitative analysis into the evaluation process. This would allow us to build more effective, transparent, and trustworthy morphing attack detection systems.
To ensure the scalability and generalizability of such systems, future work must also involve evaluating LLM-driven D-MAD pipelines on larger and more diverse public morphing datasets. Incorporating a broader range of identities, acquisition conditions, and morphing techniques would help assess the fairness, robustness, and bias resilience of these models.

\section*{ETHICAL IMPACT STATEMENT}
The primary aim of this study is to investigate the potential and limitations of multimodal Large Language Models (LLMs) in detecting face morphing attacks, with a particular emphasis on their application in secure identity verification systems such as automatic border control. This research focuses on understanding the decision-making and explainability capabilities of LLMs when analyzing facial image pairs for morphing detection.

To ensure ethical integrity, we utilize a semi-publicly available dataset in strict compliance with established data protection protocols. All data subjects included in the dataset have provided informed consent for their facial images to be used for research purposes. The study does not attempt to extract or infer private or biometric identity traits beyond what is necessary for the evaluation of morphing detection performance.

While the use of facial recognition technologies raises valid concerns regarding privacy and surveillance, this work is exclusively centered on evaluating the robustness of multimodal LLMs against image manipulation techniques (i.e., morphing), rather than deploying such models for real-world surveillance or identity inference. No personally identifiable information is disclosed, and the analysis is limited to the performance of the models on predefined tasks under controlled conditions.

By prioritizing transparency, consent, and responsible AI practices, this work seeks to promote the safe and ethical application of LLMs in high-stakes environments. Ultimately, our objective is to contribute to the development of more robust, explainable, and secure face recognition systems that can withstand adversarial attacks such as morphing, while upholding the principles of data ethics and individual privacy.

{\small
\bibliographystyle{ieee}
\bibliography{FaceMorph_Refereces}
}

\end{document}